\documentclass[review]{elsarticle}
\usepackage{tabularx}
\usepackage{lineno,hyperref}
\modulolinenumbers[5]
\usepackage{amsmath,amsfonts}
\usepackage{algorithmic}
\usepackage{algorithm}
\usepackage{array}
\usepackage{xcolor}
\usepackage[caption=false,font=normalsize,labelfont=sf,textfont=sf]{subfig}
\usepackage{textcomp}
\usepackage{stfloats}
\usepackage{soul} 
\usepackage{url}
\usepackage{verbatim}
\usepackage{graphicx}
\usepackage{tabularx,booktabs,textcomp}
\usepackage{amsthm}

\usepackage{amsmath,amssymb,amsthm}

\journal{Elsevier}

\begin{document}

\begin{frontmatter}

\title{Identification of Surface Defects on Solar PV Panels and Wind Turbine Blades using Attention based Deep Learning Model}


\author[1,2]{Divyanshi Dwivedi}

\author[1,3]{K. Victor Sam Moses Babu}

\author[2]{Pradeep Kumar Yemula}

\author[3]{Pratyush Chakraborty}

\author[1]{Mayukha Pal\corref{mycorrespondingauthor}}
\ead{mayukha.pal@in.abb.com}
\cortext[mycorrespondingauthor]{Corresponding author}


\affiliation[1]{organization={ABB Ability Innovation Center},
     addressline={Asea Brown Boveri Company},
    city={Hyderabad},
    postcode={500084},
    state={Telangana},
    country={India}}    

\affiliation[2]{organization={Department of Electrical Engineering},
    addressline={Indian Institute of Technology},
    city={Hyderabad},
    postcode={502205},
    state={Telangana},
    country={India}}

\affiliation[3]{organization={Department of Electrical and Electronics Engineering},
    addressline={Birla Institute of Technology and Science, Pilani- Hyderabad Campus},
    city={Hyderabad},
    postcode={500078},
    state={Telangana},
    country={India}}

\begin{abstract}

{The global generation of renewable energy has rapidly increased, primarily due to the installation of large-scale renewable energy power plants. However, monitoring renewable energy assets in these large plants remains challenging due to environmental factors that could result in reduced power generation, malfunctioning, and degradation of asset life. Therefore, the detection of surface defects on renewable energy assets is crucial for maintaining the performance and efficiency of these plants. This paper proposes an innovative detection framework to achieve an economical surface monitoring system for renewable energy assets. High-resolution images of the assets are captured regularly and inspected to identify surface or structural damages on solar panels and wind turbine blades. {Vision transformer (ViT), one of the latest attention-based deep learning (DL) models in computer vision, is proposed in this work to classify surface defects.} The ViT model outperforms other DL models, including MobileNet, VGG16, Xception, EfficientNetB7, and ResNet50, achieving high accuracy scores above 97\% for both wind and solar plant assets. From the results, our proposed model demonstrates its potential for monitoring and detecting damages in renewable energy assets for efficient and reliable operation of renewable power plants.}

\end{abstract}

\end{frontmatter}

\section{Introduction}
\label{section:Introduction}

\subsection{Motivation}

Renewable energy sources, particularly solar and wind power, are expected to drive a significant proportion of global power generation capacity, accounting for 75-80\% of newly installed capacity by 2050 \cite{outlook}. This shift towards green energy is primarily due to growing public demand and government policies aimed at reducing reliance on fossil fuels, achieving net-zero carbon emissions, and sustainable growth \cite{GNN}. To achieve these goals, governments around the world have prioritized planning and implementing measures to invest in large-scale renewable energy power plants. Wind and solar power have been identified by the International Energy Agency (IEA) as key sources for achieving sustainable development \cite{IEA}.

In May 2022, energy ministers from Germany, Belgium, the Netherlands, and Denmark agreed to establish a renewable energy power plant in the North Sea. This project aims to reduce dependence on Russian gas imports, thereby achieving emission reduction targets. With a projected capacity of 65 GW by 2030 and 150 GW by 2050, it represents a significant step forward in Europe's commitment to renewable energy \cite{Europe-wind}. Additionally, Europe's largest solar plant, the Núñez de Balboa, comprises 1.4 million solar panels covering almost 10 square kilometers and has an installed capacity of 500 MW \cite{Europe-solar}. Also, in Rajasthan, India, Adani Green Energy recently commissioned a solar-wind hybrid power plant with 600 MW solar and 150 MW wind capacities \cite{Adani}. These massive projects contribute to green energy but need proper management, monitoring, and maintenance.

The United Nations has identified energy management as a key element in achieving objectives for sustainable development \cite{UN}. However, the maintenance and monitoring of renewable power plants are less explored and not emphasized; they are sell-pitched as low-maintenance energy sources, which is an erroneous fact. Effective maintenance and monitoring are essential for achieving sustainable development targets, increasing power generation, and prolonging the lifespan of renewable energy assets. Without proper maintenance, renewable power plants could experience lower efficiency, increased downtime, and equipment failure.

{The blades of wind turbines are critical components that significantly impact the quality and performance of power generation \cite{ASGHAR2018495, REZAEI2016391}. However, wind turbines are often installed in remote and exposed locations which makes them vulnerable to damages caused by environmental factors such as rain, sun, and wind gusts \cite{Li2014}. Manual detection of these issues is impractical due to the requirement of a large workforce and extensive man-hours  \cite{YU20201}. An efficient, cost-effective, and reliable solution is to capture drone images and analyze them for defect detection. Meanwhile, solar PV panels are another widely used renewable energy source for small and large-scale power generation \cite{resiliency}. However, factors such as soil, dust, snow, bird droppings, construction cement deposits, cracks, and shadows from overgrown plants or grass, significantly reduce their performance and lifespan \cite{CaseStudy}. Proper maintenance of solar panels is necessary to maximize the power output throughout the lifespan of 20-25 years \cite{8679176}. Generally, to track the performance of solar panels, { the energy generation is monitored, but this is insufficient} for the identification of the root cause of reduced power generation and required preventive measures.}

\begin{figure}[H]
  \centering
  \includegraphics[width=4.6in]{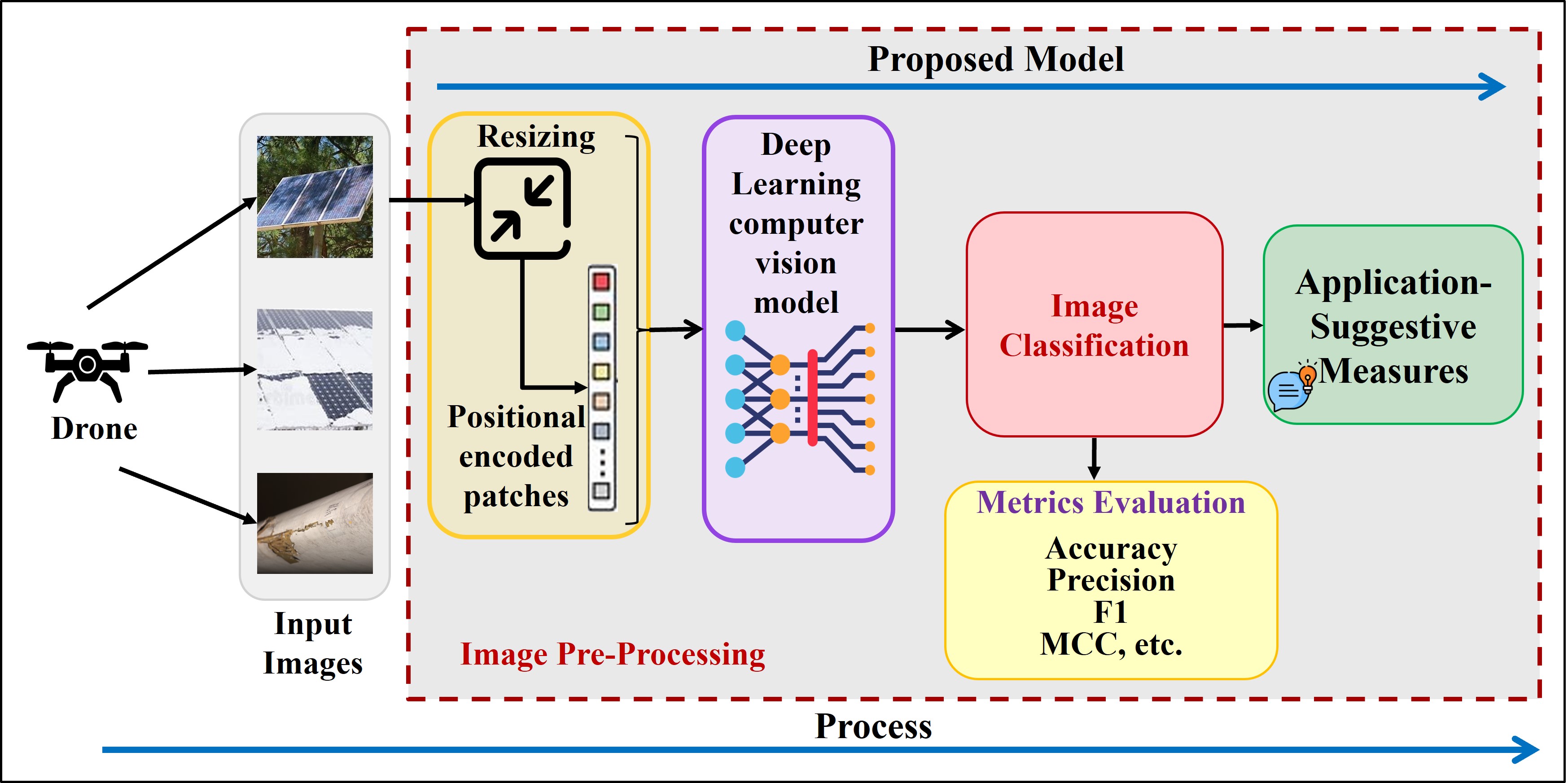}\\
  \caption{ Proposed framework for monitoring and detection of damages in the solar panels and wind turbines.}
  \label{fig:framework}
\end{figure}

This work proposes a technique for effectively monitoring and detecting damages or defects in renewable energy assets. Regular drone image-based monitoring is recommended for large-scale renewable power plants \cite{Yfantis}, and the collected images are examined to identify defects and take measures to improve power generation. Wind turbines of heights up to 65 meters and solar panels spread over 60 acres of land pose a challenge in identifying defects. Thus, the major focus is to use an automated DL-based computer vision algorithm, as depicted in Figure \ref{fig:framework}, to detect damages in wind turbines and solar PV panels deployed on a large scale. Once defects are identified, appropriate preventive measures need to be taken to enhance the performance of these assets.

\subsection{Related works}

{The use of convolutional neural networks (CNNs) has been widely adopted for the identification of defects in both solar and wind energy power plants. Pre-trained models such as VGG-16, VGG-19, Inception-v3, Inception-ResNet50-v2, ResNet50-v2, and Xception have been employed to identify micro-cracks in photovoltaic (PV) modules from electroluminescence images \cite{CNN1}. Ensemble learning has been used to improve accuracy and achieved 96.97\% and 97.06\% classification accuracy for monocrystalline and polycrystalline solar panels, respectively. { However, ensemble learning increases the complexity of the model and requires more computational resources during both training and deployment.} The application of a multi-scale SE-ResNet has been used to diagnose compound faults in PV panels covered with dust, estimating the degree of dust coverage on the PV array and the accumulation on the bottom of the PV panels \cite{LIN2022101785}. { The effectiveness of the model is sensitive to specific patterns of dust accumulation and coverage. If the model is trained on a limited range of dust conditions, its performance may degrade when exposed to diverse or unseen scenarios.} A deep CNN was applied to aerial images covering an area of 135 km\textsuperscript{2}, achieving a detection rate of 80\% for PV panels with a precision of 72\% \cite{7884415}. Deep neural networks are applied \cite{DEMIRCI2021114810} for feature extraction and machine learning methods for the classification, obtaining 90.57\% for 4 classes and 94.52\% for 2 classes. For the same dataset, semantic segmentation with CNN models is used to classify defects in PV panels, achieving an average accuracy of 70\% for 4 classes and 75\% for 2 classes \cite{RICOESPINOSA2020249}. {The obtained accuracy scores in \cite{7884415, DEMIRCI2021114810, RICOESPINOSA2020249} are significantly low, thus models are not reliable.} The implementation of six pre-trained CNN models with ImageNet has proven that the Xception model achieves higher accuracy for the classification of defective photovoltaic cells \cite{9479376}. { However, Xception models are sensitive to hyperparameter choices, and finding an optimal configuration requires careful tuning. This sensitivity usually impacts its robustness across different datasets and tasks.} CNN models have also been applied with the use of transfer learning; AlexNet was proven to be effective for identifying surface defects in PV panels \cite{9118384}. { While the model struggles to capture long-range dependencies and global context efficiently. It relies on stacking convolutional and pooling layers, which limits its ability to understand relationships between distant image regions. However, AlexNet is effective in learning hierarchical features, but it does not capture fine-grained details}. However, some models that rely solely on CNNs for visual recognition tasks have achieved a classification accuracy of 83.22\% \cite{deepeye}. Other researchers have used modified versions of CNN to detect damages in solar PV panels, but these models achieve an accuracy lower than 90\% and require more computational time to evaluate the images \cite{9492833, 9367143, SHIHAVUDDIN20214566}. 

CNN models have also been used to identify damages in wind turbine blades with an accuracy of 91\% \cite{REDDY2019106823}. Deep convolutional neural networks (DCNNs) have been employed for feature extraction with pre-trained models which is also suitable for small datasets \cite{YU20201}. In machine learning, various feature extraction techniques have been used, such as the histogram of oriented gradient (HOG) \cite{9003045}, primitive-based methods \cite{Hong1979TexturePE}, statistical methods \cite{Emrith2010MeasuringPD}, spectral methods \cite{761261}, local binary pattern (LBP) \cite{5459207}, image visibility graphs \cite{visibility1,visibility2}, and gray-level co-occurrence matrix (GLCM) \cite{RODRIGUEZGALIANO201293}. However, these typical image processing techniques only extract low-level features from the images, which may not be sufficient to identify and classify the type of defect in wind turbine blades if such models are employed. Wind turbine blades are susceptible to various types of damage, such as cracks, scrapes, and erosion, which are often non-uniform and difficult to differentiate.

In addition to the use of image processing techniques for identifying defects in renewable energy assets, other methods have been applied, such as normalized sequential voltage and current measurements from PV modules for fault diagnosis and employing a CNN model to extract features \cite{LU2019950}. {The use of normalized sequential voltage and current measurements implies a focus on temporal data without incorporating spatial information from the PV module's physical layout. This limitation may hinder the model's ability to detect spatially localized defects or anomalies that could be crucial for comprehensive fault diagnosis.} Similarly, a time-series analysis technique with a CNN model to identify faults in wind turbines is employed \cite{RAHIMILARKI2022916}. In another approach, I-V curves, temperatures, and irradiances of PV modules were analyzed under various fault conditions. A CNN model and a residual-gated recurrent unit (Res-GRU) were used to identify the PV module faults \cite{9180283}. {However, the noise, inaccuracies, or disruptions in the voltage and current measurements may impact the reliability of the CNN model's features, which could lead to false positives or negatives in fault identification.}

ViT \cite{ViT} is a new approach for image recognition that uses the transformer architecture from natural language processing (NLP). The ViT divides an image into non-overlapping patches, which are flattened and treated as sequences of tokens fed into a standard transformer model. This method outperformed various state-of-the-art models on image recognition benchmarks like ImageNet and provides greater interpretability due to the attention mechanism. ViT struggles with a lack of inductive bias and poor performance on small datasets after training; this leads to a lack of generalizability \cite{ViT}. However, pretraining on a large amount of data like ImageNet and performing transfer learning on smaller datasets help the ViT model to outperform other architectures. The application of ViT has shown significant results in biomedical science, including interpretation of chest radiography \cite{chest}, classification of oral cancer \cite{bhas1}, detection of cardiovascular disease \cite{calcium}, and many others. ViT has also proven effective for the detection of earthquakes \cite{earth1}, metal 3D printing quality recognition \cite{metal}, and for the detection of fire smoke \cite{fire}.}

\subsection{Contribution}

In this work, { an attention-based ViT model is employed} to detect damages in solar PV panels and wind turbine blades. Detecting damages from high-resolution drone images of varying modalities requires an effective model that could automatically extract high-level features from the images in a short amount of time with high accuracy. The introduction of attention to NLP in 2017 was widely appreciated for its high performance. Using this concept, Google researchers proposed a transformer-based ViT model for computer vision classification \cite{Vaswani2017AttentionIA}. In various domains, ViT has demonstrated promising performance and high accuracy for learning tasks \cite{vit1,vit2,vit3,vit4,bhas1,bhas2}. Therefore, { the transformer model is used} to identify and characterize damages on solar PV panels and wind turbine blades using the drone inspection technique.

The key contributions of this work are as follows:
\begin{itemize}
\item Implementation of ViT for the first time on an electrical power system problem.
\item The proposed framework is suitable to integrate into large-scale renewable power plants for inspecting the damage to assets at a very low cost, with less human intervention and less processing time.
\item {In comparison to other models, { the proposed model achieves} better metric scores for accuracy, recall, precision, etc., in lower execution time for the proposed model.}
\end{itemize}
The paper is organized as follows: a detailed explanation of the ViT model is presented in section \ref{section:ViT}. The pre-trained DL models used for comparison are discussed briefly in section \ref{section:pre}. The considered dataset is described in section \ref{section:Datasets}. Comparative results and analysis of the proposed ViT model with other DL models are presented in section \ref{section:Result}. Finally, section \ref{section:Conclusion} concludes the paper.

\section{Methods and materials}
\label{section:ViT}

The transformer architecture was first proposed for NLP tasks, such as machine translation, in 2017 and demonstrated outstanding performance \cite{Vaswani2017AttentionIA}. In 2021, Google's research team implemented this architecture for image processing by using the transformer encoder architecture for image recognition tasks and named it ViT \cite{ViT}. While the transformer in NLP measures the relationship between 1D input token pairs to initiate the learning process. In computer vision, images are reshaped into a sequence of flattened 2D image patches, which are used as input tokens for further learning with attention in the network. {These patches are flattened and mapped to $D_{Model}$ dimensions with a trainable linear projection to achieve a constant latent vector of size $D_{Model}$, which is used throughout the layers of the model.} This layer acts as an embedding layer and outputs fixed-sized vectors.

\begin{figure}
  \centering
  \includegraphics[width=4.5in]{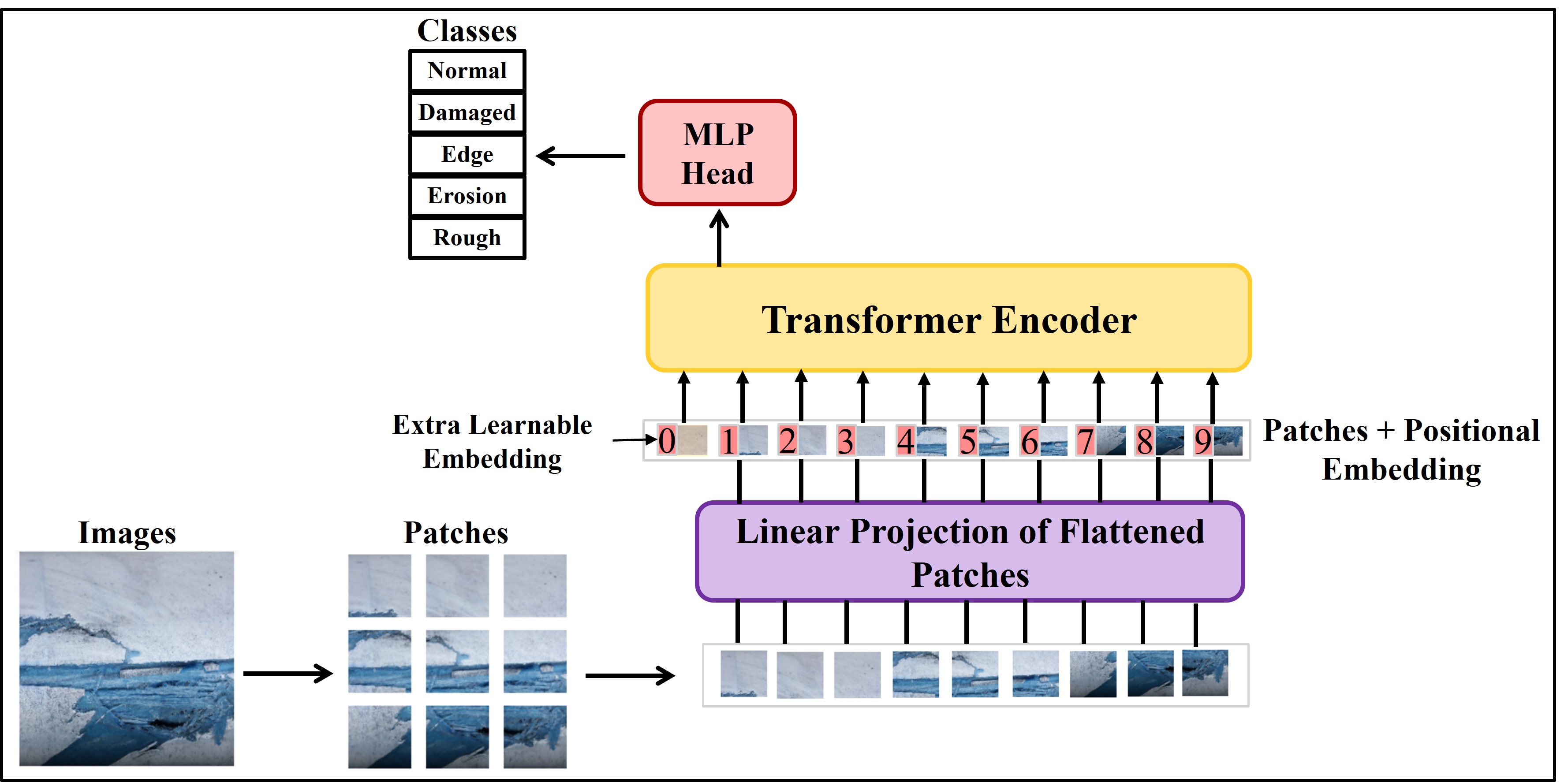}\\
  \caption{ViT framework with wind turbine blade image as input.}
  \label{fig:VitArch}
\end{figure}

Position embeddings are added linearly to the sequence of image patches so that the model retains the positional information of the images. The implementation framework of image classification using ViT is illustrated in Figure \ref{fig:VitArch}.

\subsection{Architecture of Transformer encoder}

The transformer encoder architecture block comprises stacked multi-head attention layers, a feed-forward neural network, a shortcut connection, and a normalization layer as shown in Figure \ref{fig:Attention}.

\subsubsection{Encoder stack}

{The encoder is composed of eight identical layers \cite{ViT}, each containing two sub-layers. The first sub-layer is a multi-head attention layer, while the second sub-layer is a fully connected feed-forward network layer with positional information. { The number of encoder layers was selected based on the recommendation of the original work on the ViT model \cite{ViT}; it was observed that scaling the depth results in the biggest improvements, which are visible up until 64 layers, however, diminishing returns are visible after 16 layers. Thus, we used eight identical layers for the encoder as the choice of such hyperparameters like the number of layers, and activation function in machine learning always depends upon the performance of the model.} 

A residual connection is applied between the two sub-layers, and layer normalization is performed at the end of each layer.
The output of each sub-layer is:
\begin{equation}
    LayerNorm(X+ Attention(X))
\end{equation}

where $Attention(X)$ is the function implemented by the sub-layer itself, $X$  is the input to the self-attention layer. To facilitate these residual connections, all sub-layers in the model, including the embedding layers, produce an output of dimension $D_{Model} = 512$. The outputs of the multi-head attention blocks are added and normalized using an Add and Normalization layer, allowing for residual connections to be made between layers.}

\subsubsection{Self-attention}

In the self-attention layer, the input vector is transformed into three vectors: Query Vector (Q), Key Vector (K), and Value Vector (V), each of dimension $D_Q = D_K = D_V = D_{Model} = 512$. These vectors are then stacked into their respective matrices $Q_m$, $K_m$, and $V_m$ of size $D_{Model}\times(N+1)$, where $N$ is the number of image patches { to which an extra dimension is added, i.e., a learnable (class) embedding attached to the sequence according to the position of the image patch. This extra learnable (class) embedding helps to predict the class of the input image after being updated by self-attention \cite{surveyViT}.}

Using these matrices, { the attention function is computed} as follows:

\begin{itemize}
\item {Compute the scores between the Query and Key matrices to determine the degree of attention, $S_m= Q_m \cdot\ K_m^T$.}
\item Normalize the scores for stabilizing the gradient for improving the training performance, $S_n= S_m/\sqrt{D_{Model}} $.
\item Transform the normalized scores into probabilities with the Softmax function, $P_S= Softmax(S_n)$.
\item Finally, obtain the weighted value matrix, $W_m= V_m \cdot\ P_s$.
\end{itemize}

\begin{figure}
  \centering
  \includegraphics[width=4.8in]{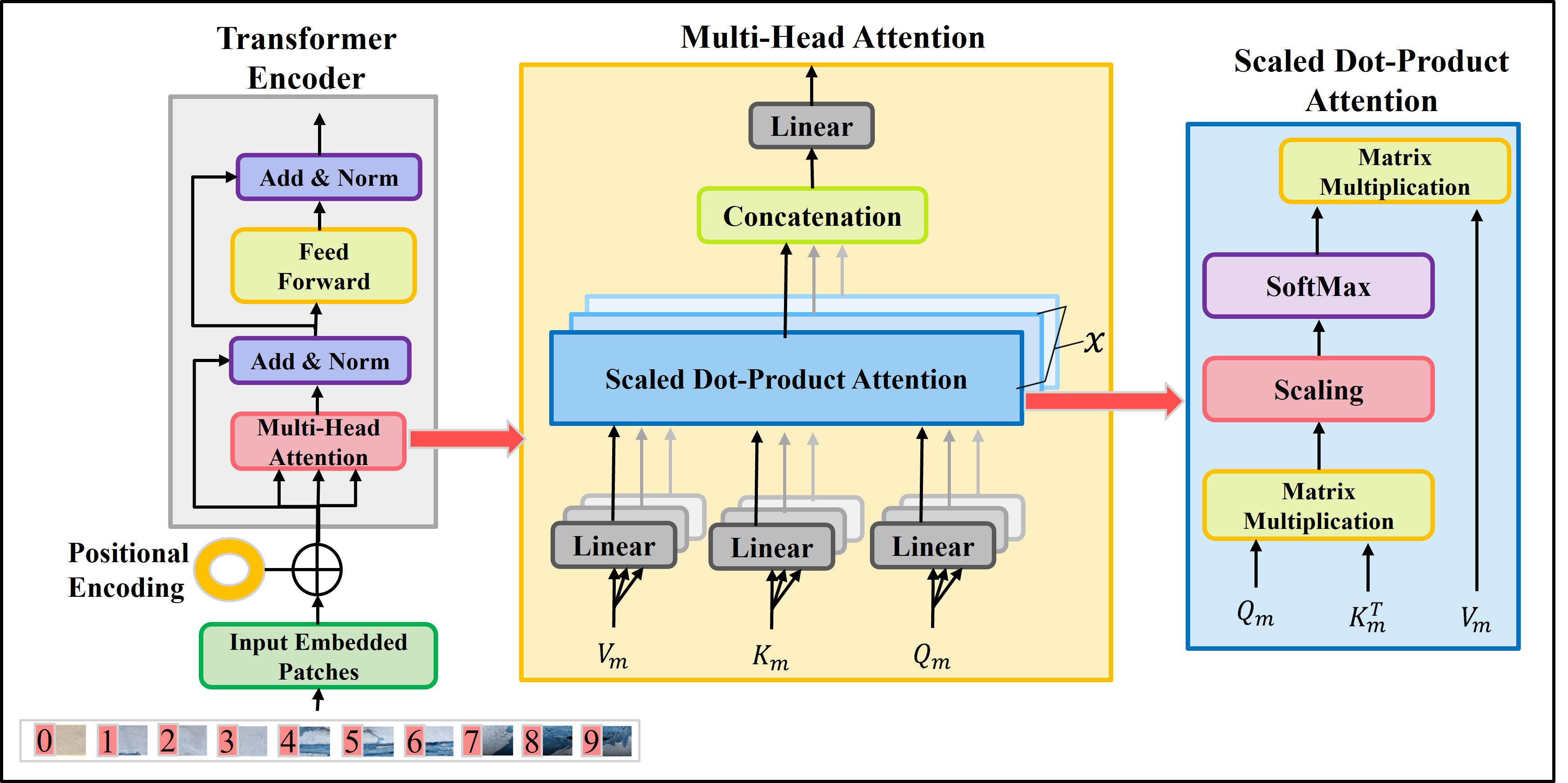}\\
  \caption{ Architectural diagram of transformer encoder with multi-head attention consists of several attention layers running in parallel and scaled dot-product attention module.}
  \label{fig:Attention}
\end{figure}

The combined expression for the self-attention function is given in equation (\ref{eq:self_atten}), and the entire process is described in Figure \ref{fig:Attention}.

\begin{equation}
   Self_{Attention(Q_m,K_m,V_m )} = \frac{Softmax(Q_m \cdot K_m^T)}{\sqrt{D_{Model}}}\cdot V_m
   \label{eq:self_atten}
\end{equation}

\subsection{Multi-head attention}

Using the single attention function $x-times$ with $D_{Model}$ dimension; the queries, keys, and values are linearly projected with different learnable linear projections to their respective dimensions. 

Further, these projected variants perform the attention function in parallel and result in $D_V$ dimensional output values as shown in Figure \ref{fig:Attention}. Multi-head attention facilitates the model to acquire complete information from the represented layers at various positions and illustrated as:
\begin{equation}
    Multi_{Head(Q_m,K_m,V_m )} = Concat(head_1,...,head_x)W^0
\end{equation}

\noindent where,$ \;head_k=Attention(Q_m W_k^{Q_m},K_m W_k^{K_m },V_m W_k^{V_m })$, and $W_k^{Q_m} \in R^{D_{Model}\times D_Q} $; $W_k^{K_m}\in R^{D_{Model}\times D_K} $; $W_k^{V_m } \in R^{D_{Model}\times D_V} $. In this work, { eight parallel attention layers or heads $(x=8)$ are used}, 
$\therefore D_Q=D_K=D_V= \frac{D_{Model}}{x} = \frac{512}{8} =64.$

\subsection{Feed-forward networks}

Each multi-head attention layer is connected to the feed-forward network as shown in Figure \ref{fig:Attention}. It is composed of two linear transformations having a ReLU activation function between them, {applied to each position separately and identically.  
\begin{equation}
    F_{FN}(x) =max(0,xW_1,b_1)\cdot W_2+b_2
\end{equation}

where, $x$ is the hidden representation at a particular position, $W_1$ and $W_2$ are two learned linear transformations matrices, and $b_1$ and $b_2$ are the bias vectors. ReLU introduces more sensitivity to weighted sum and avoids saturation. The linear projection layer helps to transform arrays into vectors while maintaining their physical dimensions \cite{ViT}.}

\subsection{Positional embedding}
Input patch images are embedded with positional encodings at the bottom of the encoder stack with the dimension $D_{Model}$. These positional encodings could be implemented by various approaches \cite{conv}, but here { sine and cosine functions at different frequencies are considered as follows,}
\begin{equation}
  PE(x,2k)=sin(x/10000^{2k/D_{model}})
\end{equation}
\begin{equation}
  PE(x,2k+1)=cos(x/10000^{2k/D_{model}})
\end{equation}

\noindent where, $x$ is the position and $k$ is the dimension. { The sine and cosine functions \cite{Vaswani2017AttentionIA} help in normalizing the values of the positional encoding matrix in the range of [-1,1]. These functions facilitate a unique way of encoding each position and quantifying the similarity between different positions.}

Thus, the process flow for implementing the ViT model is shown in Figure \ref{fig:Attention}, which includes all the layers discussed above, where linear image patches are passed through a dense layer to achieve encoded vectors by integrating them with positional embedding. The positional encoded patches are passed through transformer encoder layers to get the contextual vector. Then at the final stage, this vector is passed through a multi-layer head to get the final image classification.

{
\section{Deep learning models}
\label{section:pre}

 { The performance of the proposed ViT model is evaluated} by comparing it with five pre-trained DL models for identifying defects in solar panels and wind turbine blades. The implementation framework for these models is shown in Figure \ref{fig:DL_Model} and a brief overview of each model is given below:

\begin{itemize}

    \item MobileNet: It uses point-wise convolution and depth-wise separable convolutions \cite{mobilenet}. It has significantly fewer parameters as compared to other convolutional models with the same depth in the nets. Thus, it is considered a lightweight deep neural network.

    \item VGG16: A CNN model which is a large network comprising 138 million parameters with convolution layers of 3x3 filter with a stride 1 with same padding and max pool layer of 2x2 filter of stride 2 \cite{VGG16}. These layers are followed by two fully connected layers and softmax for output.

    \item Xception: A CNN model that consists of depth-wise separable convolution layers \cite{Xception}. Basically, it is an extreme version of the Inception model. The Inception model has 1x1 convolutions for compressing the original input; for each input space filters are applied on their respective depth space. On the other hand, Xception is performed in a reverse manner; it applies the filters on depth space and then compresses the input spaces with the help of 1x1 convolution by applying it across the depth. It does not require any non-linear function.  Xception architecture has 36 convolutional layers for feature extraction, assembled as a linear stack of residual depth-wise separable convolution layers.

    \item EfficientNetB7: It comprises the compound scaling method for uniformly scaling network width, depth, and resolution; to optimize the floating point operations per second (FLOPS) and accuracy \cite{efficient}. The model architecture has seven flipped residual blocks with different parameters and is employed with swish activation, squeeze, and excitation blocks. In the architecture, ReLU is applied to introduce non-linearity in the model.

    \item ResNet50: This CNN model is 50 layers deep that stack the residual blocks on top of each other to form a network \cite{resnet50}. It implements a concept of ``skip connection" which lies at the core of residual blocks to connect the activation layer by skipping over intermediary levels. Further, the left blocks are heaped to be used for the construction of Resnets. It enhances the model's performance by enabling regularisation to avoid the layers that affect the model's performance.
\end{itemize}

\begin{table}
\caption{ Comparison of DL models}
\begin{tabular}{>{}p{1.2cm}>{}p{0.8cm}>{}p{0.8cm}>{}p{1.3cm}>{}p{0.8cm}>{}p{4.5cm}}
\toprule
Model & \multicolumn{1}{>{}p{0.8cm}}{Refer- ence} & Size (MB) & \multicolumn{1}{>{}p{1.3cm}}{Para- meters} & \multicolumn{1}{>{}p{1cm}}{Depth} &Remarks\\
\midrule
\multicolumn{1}{>{}p{1cm}}{Mobile Net} & \cite{mobilenet} & 16 & 4.3M & 55 & \begin{tabular}[c]{@{}p{4.5cm}@{}}It is a light weight neural network. \\ It uses depthwise separable convolution layers.\end{tabular} \\
\midrule
VGG16 & \cite{VGG16} & 528 & 138.4M & 16 & It has 138 million parameters which could lead to exploding gradients problems. \\
\midrule
Xception & \cite{Xception} & 88 & 22.9M & 81 & \begin{tabular}[c]{@{}p{4.5cm}@{}}It is an extreme interpretation of the Inception model. \\It is a 71-layer deep CNN. It uses separable convolutions in depth.\end{tabular} \\
\midrule
\multicolumn{1}{>{}p{1cm}}{Efficient- NetB7} & \cite{efficient} & 256 & 66.7M & 438 & It balances the depth, width, and resolution of the network to achieve better performance. \\
\midrule
\multicolumn{1}{>{}p{1cm}}{ResNet 50} & \cite{resnet50} & 98 & 25.6M & 107 & It has a residue block to enhance the model's performance.\\
\bottomrule
\end{tabular}
\end{table}

The hyperparameters used to run the above DL models are tabulated in Table \ref{tab:hyper}.
}
\begin{figure}
  \centering
  \includegraphics[width=4.5in]{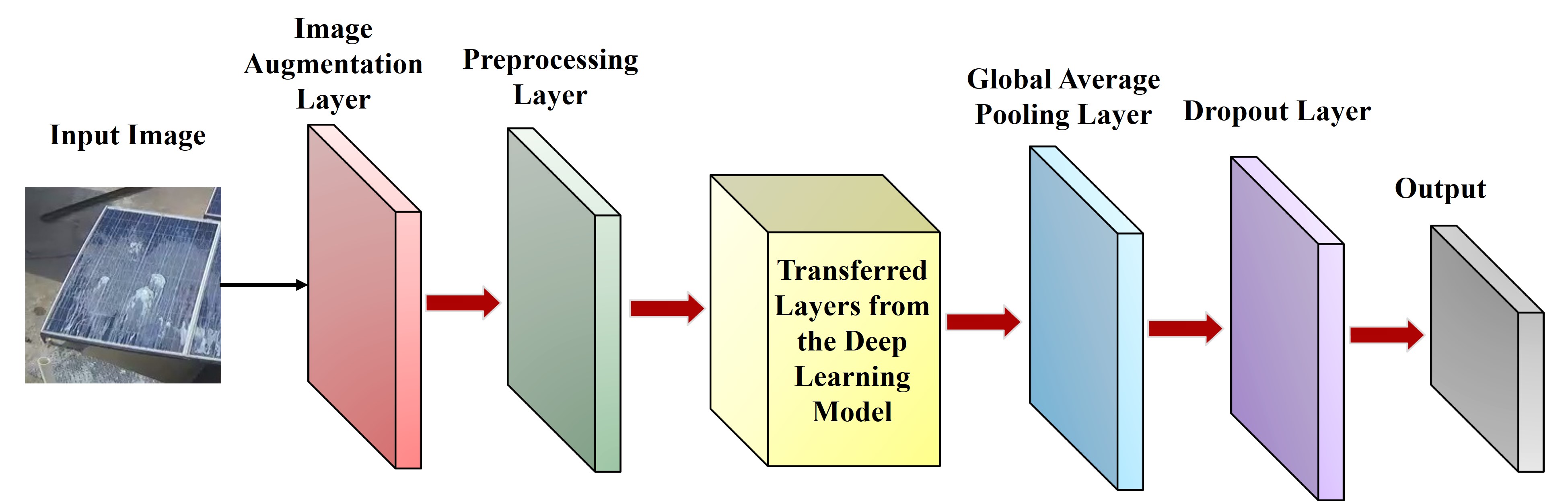}\\
  \caption{Process flow of considered DL models.}
  \label{fig:DL_Model}
\end{figure}

\section{Description of datasets}
\label{section:Datasets}
\subsection{Wind turbine blades dataset}
\begin{figure}[H]
  \centering
  \includegraphics[width=4in]{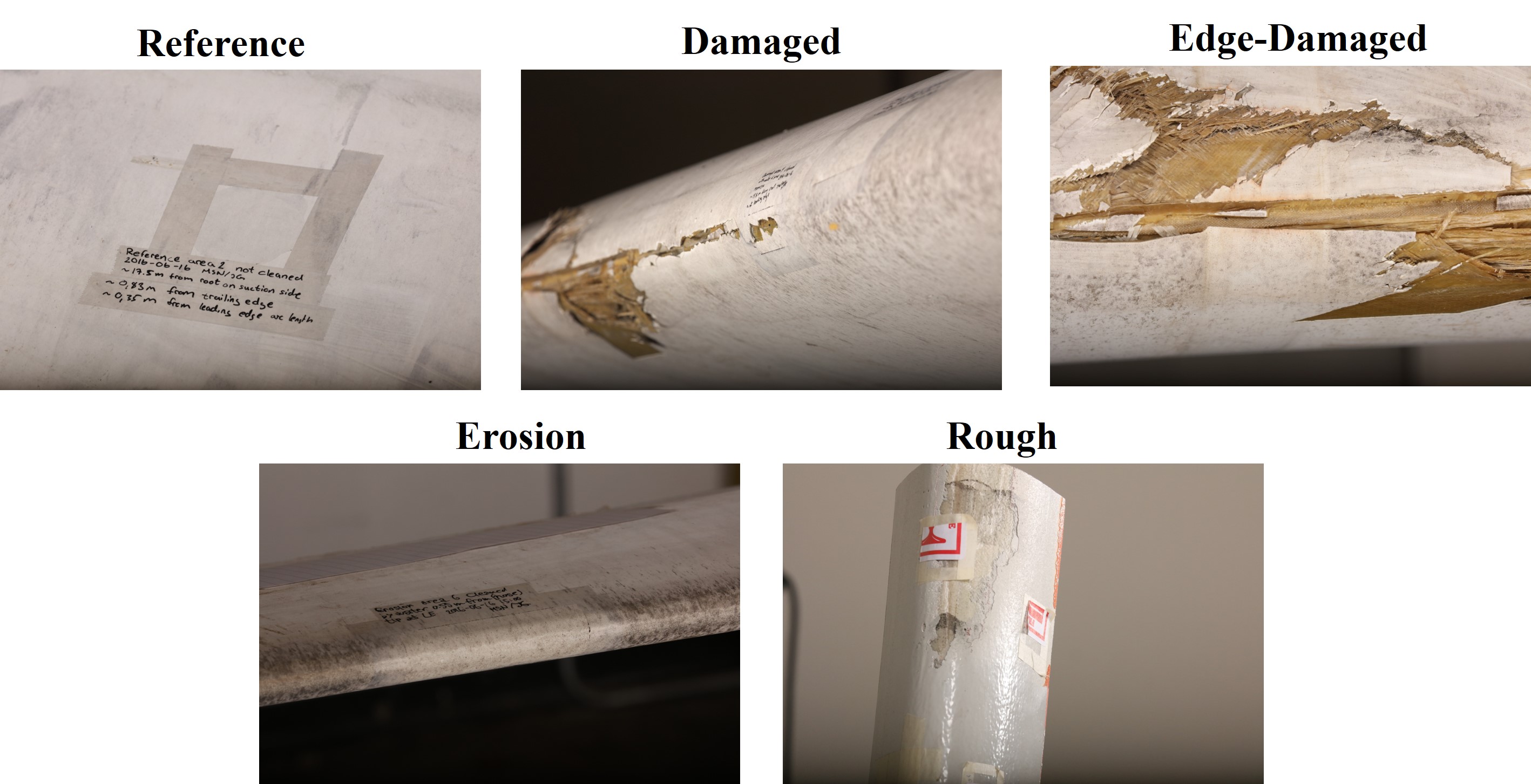}\\
  \caption{Surface damages on wind turbine blades.}
  \label{fig:wind}
\end{figure}

For the identification of defects on the blades of the wind turbine, the dataset is taken from Mendeley-Drone inspection images of a wind turbine \cite{Nikolov2020WindTB}. The images of wind turbine blades are captured using a Canon 5Ds DSLR camera with a resolution of 8688$\times$5792. The dataset comprises of images of wind turbine blades without any defects which are considered as reference images, as well as images with defects including damaged area, damaged-edge area, erosion area, and space of rough area as shown in Figure \ref{fig:wind}. In total, there are 299 images which are further labeled into five classes for training the image processing model as shown in Table \ref{tab:windstat}.

\begin{table}[H]
  \centering
  \caption{Description of wind turbine blade images}
\begin{tabular}{cc}
\toprule
Type of Image & Number of Images \\
 \midrule
Reference     & 16               \\
Damaged       & 30               \\
Edge-Damaged  & 58               \\
Erosion       & 65               \\
Rough         & 130             \\
\midrule
Total         &  299            \\
\bottomrule
\end{tabular}
\label{tab:windstat}
\end{table}

\subsection{Solar panels dataset}
\begin{figure}[H]
  \centering
  \includegraphics[width=4in]{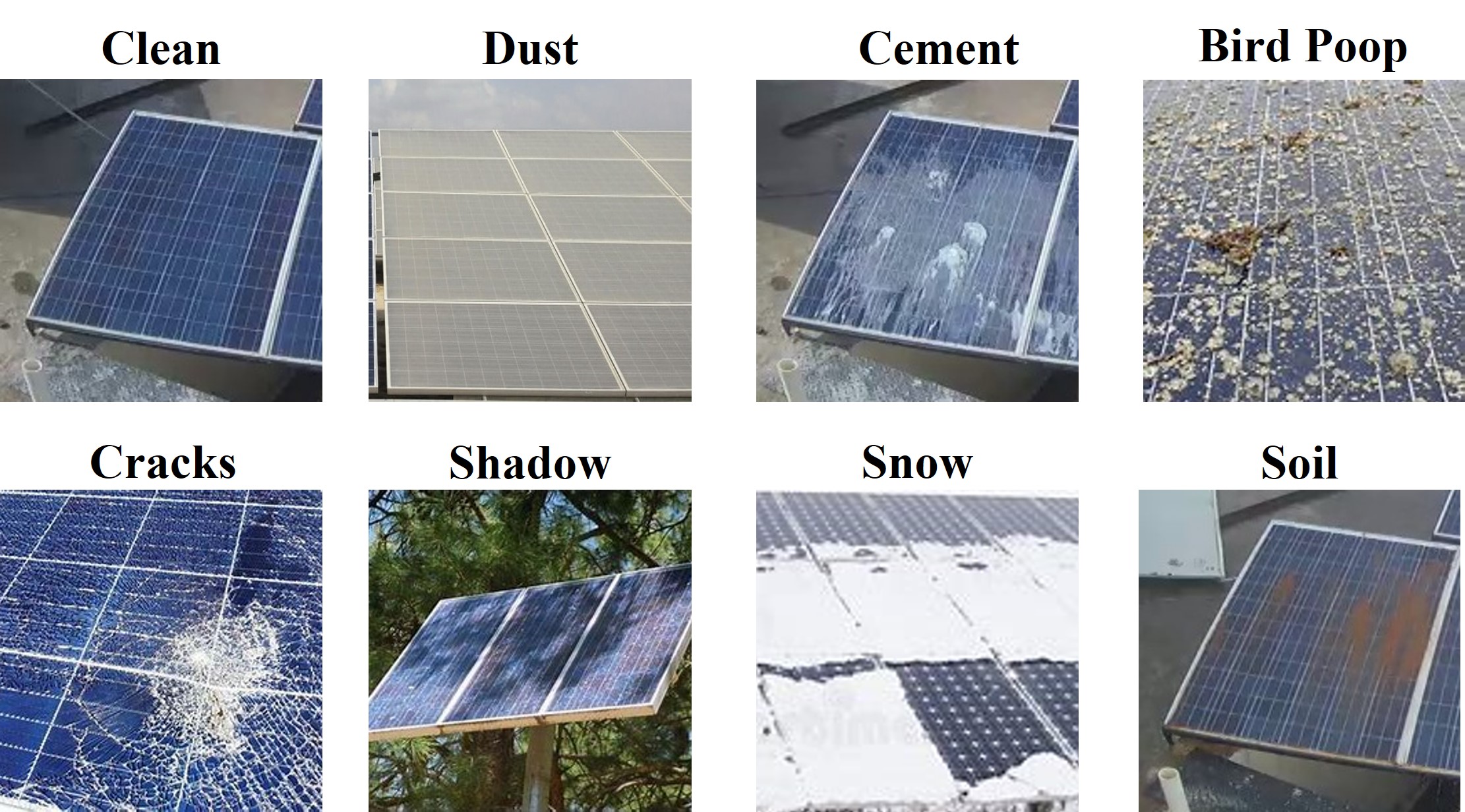}\\
  \caption{Surface damages on solar panels.}
  \label{fig:solar}
\end{figure}

To identify defects in solar panels, { the solar panel soiling image dataset created by deep solar eye \cite{deepeye} is used}. This dataset contains a total of 45,469 images captured by an RGB camera every 5 seconds for a month, with a resolution of 192$\times$192. The images were captured under various fabricated circumstances, such as sand, dust, soil, and white powder, to demonstrate the impact of soiling on the solar panels. In addition, { Google images of solar panels are used} with other types of defects, including bird droppings or nests, snow coverage, cracks, shadows from trees, plants, or buildings, and hardened cement, as shown in Figure \ref{fig:solar}. These images were resized to a resolution of 72$\times$72 for further image processing. { the images are labeled} into different classes based on the type of defect, and the number of images in each class as shown in Table \ref{tab:solarstat}.

\begin{table}[H]
\centering
  \caption{Description of solar panel images}
\begin{tabular}{cc}
\toprule
Type of Image & Number of Images \\
\midrule
Clean         & 267              \\
Dust          & 1204             \\
Cement        & 760              \\
Bird Droppings     & 165              \\
Cracks        & 73               \\
Snow          & 605              \\
Soil          & 980              \\
Shadow        & 56              \\
\midrule
Total         &  4110            \\
\bottomrule
\end{tabular}
\label{tab:solarstat}
\end{table}

{
\subsection{Image augmentation}

In our analysis, the image data is not large enough to achieve effective generalization by training our DL models on the original data alone. To address this, { a technique referred to as image augmentation \cite{bhas1} is used} to generate new images for training. { The ImageDataGenerator function from Keras DL toolbox is used} to generate sets of tensor image data with relevant data augmentation. The ImageDataGenerator takes a batch of images and applies techniques such as random rotation, flip, shift, standardization, formatting, and zoom to each image in the batch. Table \ref{tab:aug_hyper} shows the parameters used for augmentation in comparative DL models and ViT model. For example, ``Random Flip= Horizontal" indicates that images are flipped horizontally. ``Random Height= 0.2" means that images are shifted upward or downward by a factor of 0.2. ``Random zoom = 0.2$\times$0.2" zooms the images by a height factor of 0.2 and a width factor of 0.2.
}

\begin{table}
  \centering
  \caption{ Parameters of image augmentation for DL models and the proposed ViT model}
\begin{tabular}{>{}c>{}c>{}c}
 \toprule
Random Parameters      & DL models & ViT model \\
   \midrule
Rotation         & 0.02  & 0.02                                        \\
Flip   & Horizontal     & Horizontal                                      \\
Weight           & 0.2 & -                                         \\
Height     & 0.2  & -                                        \\
Zoom      & - & 0.2$\times$0.2 \\

    \bottomrule
\end{tabular}
\label{tab:aug_hyper}
\end{table}

\section{Results and discussion}
\label{section:Result}

{ All the analyses was performed on Python 3.7.6, TensorFlow 2.7.0, and Keras 2.7.0 on a standard PC with Intel(R) UHD Graphics 620. The processor is Intel(R) Core(TM) i5-8365U CPU processor @ 1.60GHz (8 CPUs), 16.0 GB of RAM, and the operating system is Windows 10 Enterprise 64-bit. For training, { 100 epochs are considered in each run}. The dataset is split into 75\% for training and the remaining 25\% for validation.} In this section, { a discussion is provided on the various evaluation metrics that are used} to analyze the performance of the proposed model and the comparison with the considered DL models on the solar panels and wind turbine blades image datasets.

\subsection{Metrices for model analyses}
\label{metrices}
{To analyze the performance of the model, four possible categories for labeling are computed which include true positive (TP), true negative (TN), false positive (FP), and false negative (FN). Here, true (T) and false (F) represent that the model has correctly and incorrectly classified the images respectively. TP and TN  depict that the model has correctly classified into the positive and negative classes, respectively. FP  means the model classified an observation to be positive when in reality, it was actually negative. FN means the model incorrectly classified an observation as negative when it should have been classified as positive.} Using these categories of classification, the following scores are computed:
\begin{itemize}
    \item Accuracy- It is widely used to analyze the model effectiveness, which compares the total number of accurate predictions concerning the total number of guesses.
    \begin{equation}
        Accuracy_{Score}=  \frac{TP+TN}{(TP+FN+FP+TN)}
    \end{equation}
    
    \item Recall- It measures the success of prediction under misbalancing. Mathematically, the ratio between truly classified positive cases to the sum of TP and FN is defined as:
    \begin{equation}
        Recall_{Score}=  \frac{TP}{(TP+FN)}
    \end{equation}
    
    \item Precision- It measures the model's ability to not label the positive sample as negative. Mathematically, it is expressed as the ratio of true positive to the total predicted positive defined as: 
    \begin{equation}
       Precision_{Score}=  \frac{TP}{(TP+FP)}
    \end{equation}

    \item F1 Score- It is a harmonic mean of precision and recall scores \cite{7055841} as:
     \begin{equation}
       F1_{Score}=  \frac{2 \quad Precision_{Score} \times Recall_{Score}}{(Recall_{Score}+Precision_{Score})}
    \end{equation}
    
    \item Cohen's Kappa- It is a score that measures inter-annotator agreement, which tells how effective the classifier model is performing compared to the classifier that randomly performs the classification. Mathematically, it is expressed as:
\begin{equation}
   Cohen\_ Kappa_{Score}= \frac{P_o-P_e}{1-P_e} 
\end{equation}
    
where, $P_o$ is the observed classification and $P_e$ is the expected classification.
    \item Matthews Correlation Coefficient (MCC)- It is the most effective and truthful score to evaluate any classifier model. Mathematically, it is expressed as:
\begin{multline}
    Matthew\_Corr_{Score}= \\ \frac{(TP \times TN-FP \times FN)}{\sqrt{(TP+FP)(TP+FN)(TN+FP)(TN+FN)}}
\end{multline}
    
\end{itemize}
{
\subsection{Hyperparameter tuning}
The solar panel images are resized to 72$\times$72, and the wind turbine blade images are resized to 256$\times$256 for the considered DL models and the proposed ViT model. In the solar dataset, the size of the original images are varied and less than the optimal size of 256$\times$256. {A few images from Google are also included} that are also varied in size and lower than 256$\times$256. Therefore, to have uniform dimensions, { all images in the dataset are resized to 72$\times$72}. In the wind turbine dataset, the original image size is 8688$\times$5792, which is quite large. In order to reduce the computation time, { the images are resized to the optimal size of 256$\times$256}. In the proposed ViT model, the performance is not affected due to image resizing. The resizing is preferred to bring uniformity in the image dataset as the images are of different sizes. It also helps in reducing the computation and training time \cite{resize}. Other hyperparameters are adjusted to get the generalized performance of the models.

\subsubsection{Hyperparameters for proposed ViT model}

{ Sparse categorical cross entropy is selected} as a loss function for multi-class classification, where the output label assigns an integer value $(0, 1, 2, \dots)$. {AdamW optimizer is used} for optimization, a stochastic gradient descent method based on adaptive estimation of first-order and second-order moments with an added method to decay weights of 0.0001. The learning rate is taken to 0.001 to achieve a minimum loss function for 100 epochs with a batch size of 32. In the encoder architecture, {a dropout of 0.5, 8 heads, and 8 transformer layers is selected.}

\subsubsection{Hyperparameters for considered DL models}

The best-suited hyperparameters are chosen for DL models; categorical cross entropy is used as a loss function for multi-class classification so that the target variable takes multiple values. Then, for optimization, { Adam is used}, which is a combination of the gradient descent with the momentum algorithm and the root mean square propagation algorithm. {A total of 100 epochs are considered} with a batch size of 32 and a learning rate of 0.001. To avoid overfitting, {a dropout of 0.2 is selected}. All the hyperparameters used are shown in Table \ref{tab:hyper}.}

\begin{table}
\caption{ Hyperparameters for DL models and the proposed ViT model}
\begin{tabular}{>{}p{3cm} >{}p{3cm} >{}p{3cm}}
 \toprule
Hyperparameters      & \multicolumn{1}{>{}p{3cm}}{DL models} & ViT model \\
\midrule
Batch   Size         & 32  & 32                                         \\
Number   of Epochs   & 100     & 100                                     \\
Optimizer            & Adam & AdamW                                         \\
Learning   Rate     & 0.001  & 0.001                                        \\
Loss Function      & \multicolumn{1}{>{}p{3cm}}{Categorical Crossentropy} & 
\multicolumn{1}{>{}p{3cm}}{Sparse Categorical Crossentropy} \\
Weight   Decay       & - & 0.0001                                       \\
Dropout             & 0.2 & 0.5 \\
Transformer   Layers & - & 8                                            \\
Heads                & - & 8                                            \\
Project   Dimension  & - & 64                                           \\
Patch   Size       & -  & Wind   Turbine Blades- 16, Solar Panels- 8  \\
    \bottomrule
\end{tabular}
\label{tab:hyper}
\end{table}

\begin{figure}[H]
  \centering
  \includegraphics[width=4.5in]{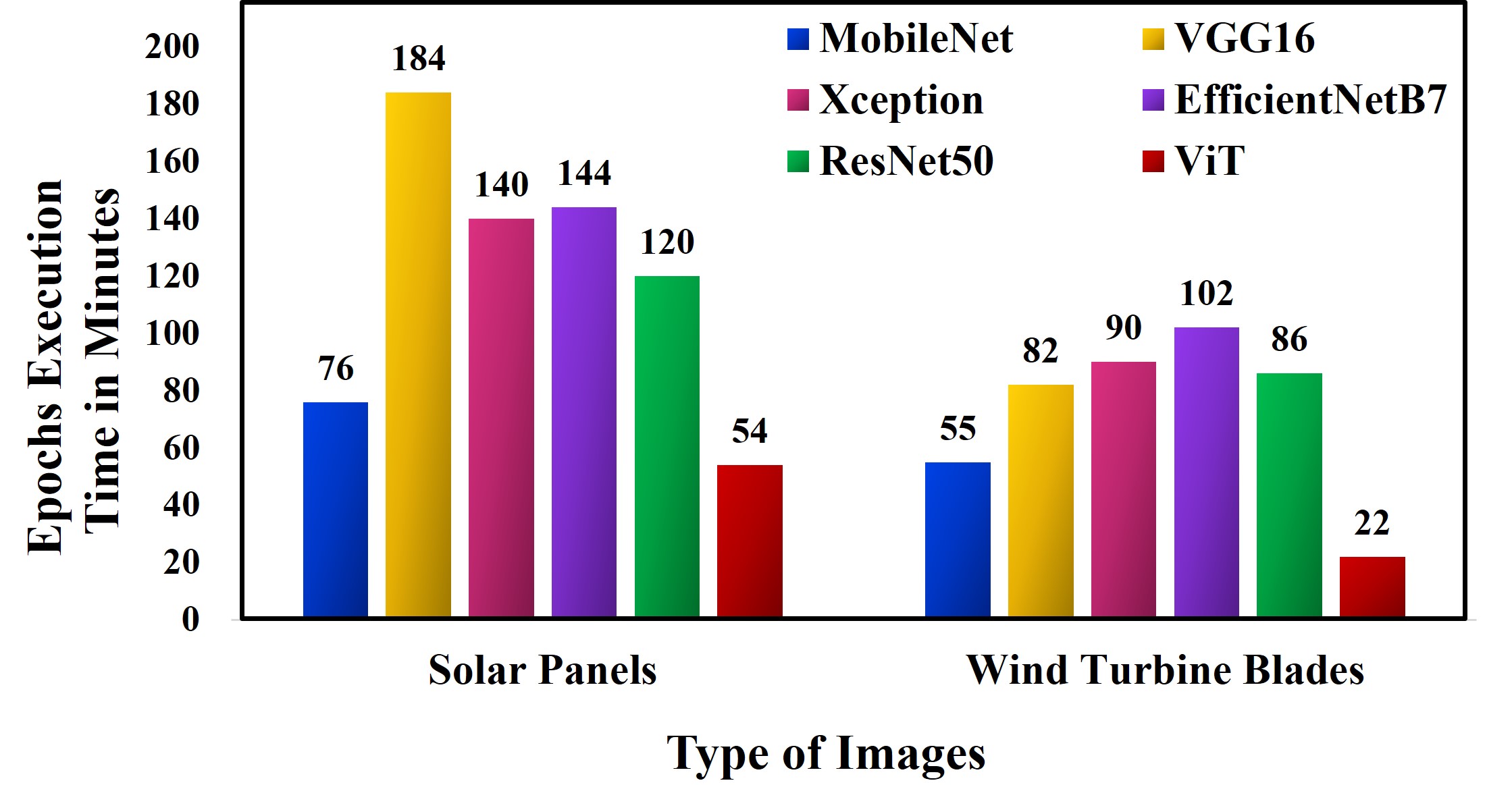}\\
  \caption{ Time required for executing image classification using DL models and proposed ViT model in minutes.}
  \label{fig:time}
\end{figure}

{

\begin{figure}
  \centering
  \includegraphics[width=3in]{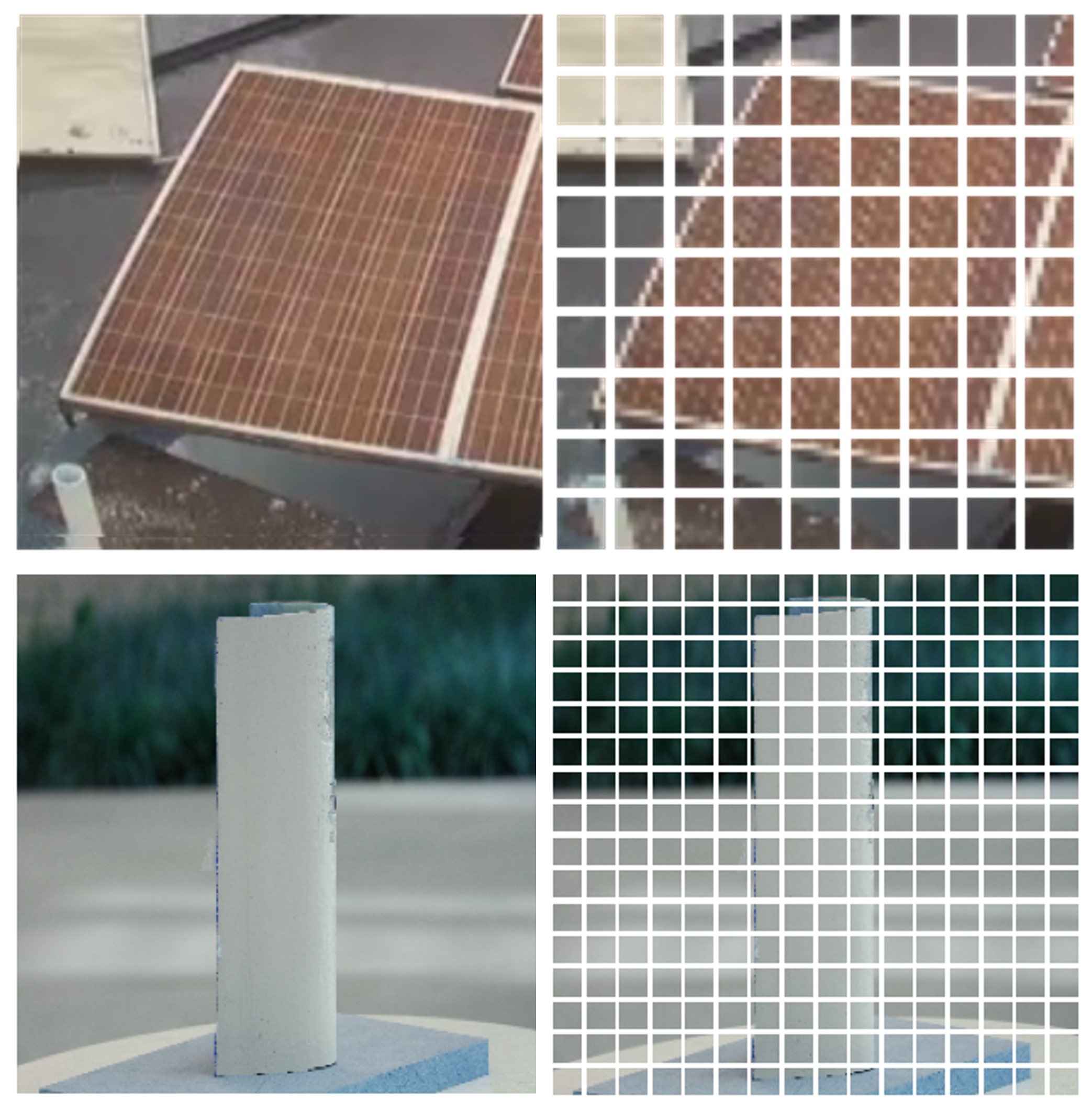}\\
  \caption{Solar panel image of size 72$\times$72 divided into 81 patches of size 8$\times$8 (top), {wind turbine blade image of size 256$\times$256 divided into 256 patches of size 16$\times$16 (bottom).}}
  \label{fig:patch}
\end{figure}
{

\subsection{Results and discussion for ViT model in comparison to other DL models}

In the ViT model, the resized images of wind turbine blades and solar panels are flattened into 2D image patches of size 16$\times$16 and 8$\times$8, respectively, as shown in Figure \ref{fig:patch}. { A pre-trained ViT model is used} on the ImageNet dataset. The accuracy and cross-entropy curves with 100 epochs for training and validation are shown in Figure \ref{fig:accuracy} for solar panels and wind turbine blades for the proposed ViT model. For testing the dataset on the trained model, { a confusion matrix is computed} to capture the number of TP, FP, TN, and FN as shown in Figure \ref{fig:confusion}. From the confusion matrix, { it is inferred} that the number of FP and FN are very low in comparison to the TN and TP, which shows that the images are classified correctly. Further, to analyze how effectively each defect is identified and classified, { a barplot is shown in Figure \ref{fig:report}} for all the classes: cement, dust, snow, and soil in the solar panel's dataset are classified with higher values of precision, recall, and F1 score; thus they are highly sensitive. The metric scores are less in the case of shadow, crack, and bird droppings because of the availability of a small number of images. For wind turbine blade images, reference and edge defects are classified correctly, but precision for damage is a little less in comparison to other classes. The performance of the model is compared and analyzed based on the metrics explained in Section \ref{metrices}. }

\begin{figure}
  \centering
  \includegraphics[width=4.5in]{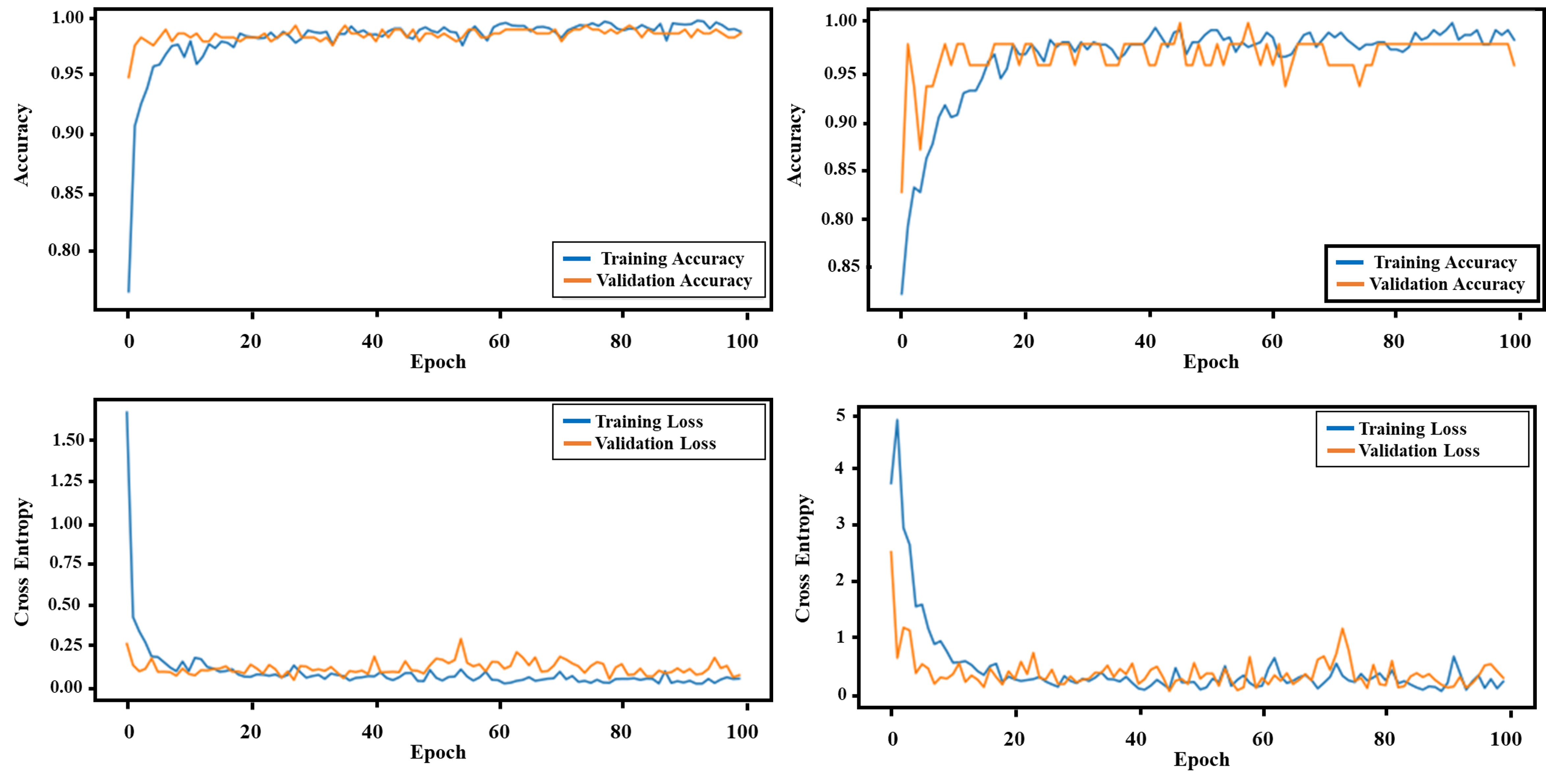}
  \caption{Accuracy and loss curve with respect to epochs of training and validation for solar panels images (left), and wind turbine blades images (right).}
  \label{fig:accuracy}
\end{figure}

\begin{figure}
  \centering
  \includegraphics[width=4.5in]{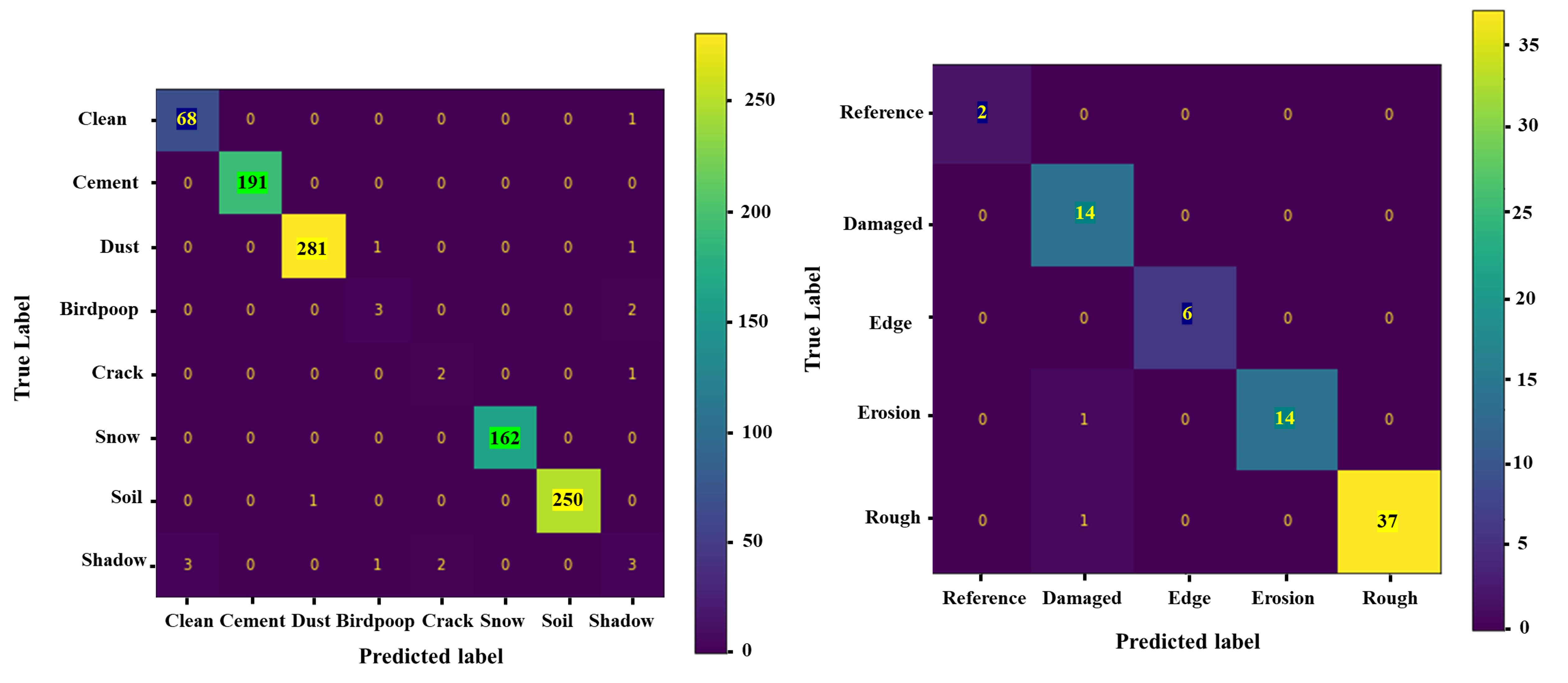}
  \caption{Confusion matrix showing all the labels for (left) solar panel images and (right)  wind turbine blade images.}
  \label{fig:confusion}
\end{figure}

\begin{figure}
  \centering
  \includegraphics[width=4.5in]{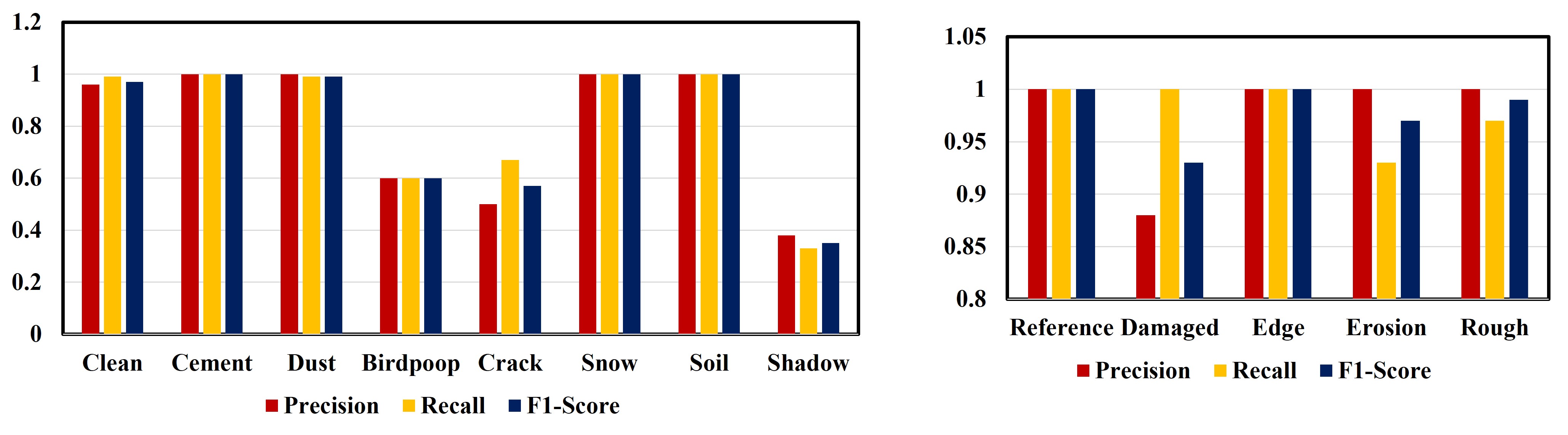}
  \caption{Reports precision, recall, and F1 scores for each class labels for (left) solar panels images, and (right)  wind turbine blade images.}
  \label{fig:report}
\end{figure}

\subsubsection{Results and discussion for solar panels dataset}

For analytical comparison, { all the pre-trained DL models, as well as the proposed ViT model, are applied} on the solar panels dataset and computed the metrics to evaluate the performance of the models. { It is} observed in Table \ref{tab:scores_sp}, for the MobileNet model, the metric scores are significantly low in comparison to other models, but its execution time is low because of the requirement of fewer parameters, as shown in Figure \ref{fig:time}. However, the proposed ViT model outperforms in terms of metric scores as well as execution time in comparison to the other models. The highest accuracy is obtained for the ViT model, i.e., 98.66\%, which is a measure of how correctly defects are classified. However, accuracy could mislead the performance of the models. Thus, both the recall and precision scores were computed to be 0.990 for the ViT model, which is very high in comparison to other DL models, and this shows how correctly images are positively labeled. To balance the recall and precision score, { the harmonic mean is computed} for computing the F1 score and achieved a high value of 0.9950 for the ViT model. However, it is observed that other DL models could mislead the detection of defects as recall, precision, and F1 scores are comparably low. Further, { Cohen's Kappa score is computed}, which is preferred to be close to 1, depicts that the predicted labels are correct and do not have randomness. The obtained value is 0.9828, which is significantly higher. The MCC is a robust and reliable metric. It would produce a high score only if the prediction obtained good results in all four categories of the confusion matrix. The MCC score obtained is 0.9828 for the ViT model. From Figure \ref{fig:time}, it is observed that for the execution of 100 epochs, the proposed ViT model requires approximately 1 hour to train the model, which is a lower training time required compared to other models.

\subsubsection{Results and discussion for wind turbine blades dataset}

Similarly, { pre-trained DL models, as well as the proposed ViT model, are implemented} on the wind turbine blades dataset and evaluated the metric scores to analyze the performance of models used for detecting defects on the surface of wind turbine blades. These metric scores are less in comparison to the solar panels because the number of images available for training is less, as shown in Table \ref{tab:scores_wbt}. The DL models could not give an accuracy of more than 94\%. On the other hand, ViT gives 97.33\% accuracy for detecting the defects and classifying them correctly. { The recall, precision, and F1 scores are computed} for the wind turbine blades dataset and obtained better score values in comparison to the other DL models. Also, the values of the MCC score are low for all the considered DL models when compared to ViT; { MCC score of 0.9635 is achieved}. When analyzing the time of execution required to run 100 epochs by DL models and the proposed ViT model from Figure \ref{fig:time}, it is observed that the ViT model just took 22 minutes to train the model. From the above results, it is proven that the proposed ViT model based on the attention mechanism is effective and superior when compared to other DL models.}

\begin{table}[H]
\centering
\centering\caption{Performance evaluation metrics for solar panel image classification model}
\begin{tabular}{>{}p{2cm} >{}p{1cm} >{}p{1cm} >{}p{1cm} >{}p{1cm} >{}p{1cm} >{}p{1cm}}
\toprule
Scores    & \multicolumn{1}{>{}p{1cm}}{Mobile Net}	& VGG16	 & \multicolumn{1}{>{}p{1cm}}{Xcept- ion}	& \multicolumn{1}{>{}p{1cm}}{Efficient NetB7}	& \multicolumn{1}{>{}p{1cm}}{ResNet 50}	& ViT\\
\midrule
Accuracy  & 0.8829	& 0.9211	& 0.9301	& 0.9455	& 0.9658 & 0.9866       \\
Recall     & 0.8903	& 0.9267	& 0.9428	& 0.9489	& 0.9682 & 0.9900     \\
Precision       & 0.8900	& 0.9224	& 0.9428	& 0.9489	& 0.9685 & 0.9900     \\
F1    & 0.8901	& 0.9245	& 0.9428	& 0.9489	& 0.9683	& 0.9950 \\
Cohen’s Kappa     & 0.7849	& 0.8512	& 0.8712	& 0.9098	& 0.9388	& 0.9828 \\
MCC & 0.7849	& 0.8530	& 0.8712	& 0.9097	& 0.9398	& 0.9828  \\
\bottomrule
\end{tabular}
\label{tab:scores_sp}
\end{table}

\begin{table}[H]
\centering\caption{Performance evaluation metrics for wind turbine blade image classification model }
\begin{tabular}{>{}p{2cm}>{}p{1cm}>{}p{1cm}>{}p{1cm}>{}p{1cm}>{}p{1cm}>{}p{1cm}}
\toprule
Scores & \multicolumn{1}{>{}p{1cm}}{Mobile Net}	& VGG16	 & \multicolumn{1}{>{}p{1cm}}{Xcept- ion}	& \multicolumn{1}{>{}p{1cm}}{Efficient NetB7}	& \multicolumn{1}{>{}p{1cm}}{ResNet 50}	& ViT \\
\midrule
Accuracy   & 0.8669	& 0.8729	& 0.9020	& 0.9127	& 0.9408	& 0.9733       \\
Recall  & 0.8702	& 0.8826	& 0.9085	& 0.9224	& 0.9488	& 0.9754 \\
Precision    & 0.8702	& 0.8838	& 0.9088	& 0.9248	& 0.9454	& 0.9829 \\
F1   & 0.8702	& 0.8832	& 0.9086	& 0.9236	& 0.9471	& 0.9791 \\
Cohen’s Kappa    & 0.7336	& 0.7468	& 0.8422	& 0.8562	& 0.8828	& 0.9627 \\
MCC & 0.7364	& 0.7482	& 0.8458	& 0.8698	& 0.8828	& 0.9635 \\
\bottomrule
\end{tabular}
\label{tab:scores_wbt}
\end{table}

\vspace{1cm}

\section{Conclusion}
\label{section:Conclusion}
{ To ensure high power generation and low maintenance costs for renewable energy assets; regular monitoring and defects detection of drone-inspected images is important. In this paper, { the surface defects are identified} from the images of wind turbine blades and solar panels by performing multi-class image classification using an attention-based ViT model. The results showed that the ViT model has effectively classified the damages in solar panels and wind turbine blades with an accuracy of 98.66\% and 97.33\% and MCC scores of 0.9829 and 0.9635, respectively. The ViT model also outperformed other DL models like MobileNet, VGG16, Xception, EfficientNetB7, and ResNet50 in terms of metric scores and computational time.

The application of the proposed model could also be used for detecting incipient faults. Generally, these faults pose challenges as the signatures are less evident due to low magnitude. However, a separate class with a set of sufficient training images associated with incipient faults could be created and the proposed model be trained for detection and effective class prediction for such defects. Thus, the attention-based ViT model for inspecting renewable energy assets would enhance the life span, reduce the maintenance cost, generate more power, and provide information to take corrective measures appropriately. This model would come out as an early intelligent system to monitor and detect the structural damages on the surface of renewable energy assets of the large-scale power utilities.}

\section*{Acknowledgement}

Divyanshi Dwivedi and K. Victor Sam Moses Babu would like to thank ABB Ability Innovation Centre, Hyderabad for the financial support in research. The author Mayukha Pal would like to thank the ABB Ability Innovation Center, Hyderabad, for their support in this work.

\section*{CRediT authorship contribution statement}

\textbf{Divyanshi Dwivedi:} Methodology, Software, Data curation, Writing- Original draft. \textbf{K. Victor Sam Moses Babu:} Methodology, Data curation, Writing- Original draft. \textbf{Pradeep Kumar Yemula:} Supervision, Writing- Reviewing and Editing. \textbf{Pratyush Chakraborty:} Supervision, Writing- Reviewing and Editing. \textbf{Mayukha Pal:} Conceptualization, Methodology, Project administration, Validation, Supervision, Writing- Reviewing and Editing.

\section*{Declaration of Competing Interest}

The authors declare that they have no known competing financial interests or personal relationships that could have appeared to influence the work reported in this paper.

\bibliography{revised_R2}

\end{document}